\documentclass[10pt]{article} 
\usepackage[accepted]{tmlr}

\usepackage{amsmath,amsfonts,bm}

\def\eqref#1{equation~\ref{#1}}

\def\ceil#1{\lceil #1 \rceil}
\def\floor#1{\lfloor #1 \rfloor}
\def\1{\bm{1}}

\DeclareMathAlphabet{\mathsfit}{\encodingdefault}{\sfdefault}{m}{sl}
\SetMathAlphabet{\mathsfit}{bold}{\encodingdefault}{\sfdefault}{bx}{n}

\usepackage{hyperref}
\usepackage{url}
\usepackage{xcolor}
\usepackage{amsmath}
\usepackage{amsfonts}
\usepackage{amssymb}
\usepackage{algpseudocode}
\usepackage{bbm}
\usepackage{stmaryrd}
\usepackage{xspace}
\usepackage{float}
\usepackage{array} 
\usepackage{textcomp}
\usepackage{graphicx}
\usepackage[linesnumbered,ruled,vlined]{algorithm2e}
\usepackage{longtable}
\usepackage{subcaption}

\title{Fuzzy PyTorch: Rapid Numerical Variability Evaluation for Deep Learning Models}

\author{\name Inés Gonzalez-Pepe \email i\_gon@live.concordia.ca \\
  \addr Department of Computer Science and Software Engineering\\
  Concordia University \\
  Montreal, Canada \AND
  \name Hiba Akhaddar \email h\_akhadd@live.concordia.ca \\
  \addr Department of Computer Science and Software Engineering\\
  Concordia University \\
  Montreal, Canada \AND \name Tristan Glatard \email
  tristan.glatard@camh.ca \\
  \addr Krembil Centre for Neuroinformatics \\
  Centre for Addiction and Mental Health\\
  Toronto, Canada \AND \name Yohan Chatelain \email
  yohan.chatelain@camh.com \\
  \addr Krembil Centre for Neuroinformatics \\
  Centre for Addiction and Mental Health\\
  Toronto, Canada}
  
\def\month{01}  
\def\year{2026} 
\def\openreview{\url{https://openreview.net/forum?id=0ogq232VGP&noteId=BtOp1tw1dN}} 

\begin{document}

\maketitle

\begin{abstract}
We introduce Fuzzy PyTorch, a framework for rapid evaluation of numerical variability in deep learning (DL) models. As DL is increasingly applied to diverse tasks, understanding variability from floating-point arithmetic is essential to ensure robust and reliable performance. Tools assessing such variability must be scalable, efficient, and integrate seamlessly with existing frameworks while minimizing code modifications. Fuzzy PyTorch enables this by integrating stochastic arithmetic into PyTorch through Probabilistic Rounding with Instruction Set Management, a novel library interfacing with Verificarlo, a numerical analysis compiler. The library offers stochastic rounding mode and a novel mode; up-down rounding.
Comparative evaluations show Fuzzy PyTorch maintains model performance and achieves runtime reductions of $5\times$ to $60\times$ versus Verrou, a state-of-the-art tool. We further demonstrate scalability by running models from 1 to 341 million parameters, confirming applicability across small and large DL architectures. Overall, Fuzzy PyTorch provides an efficient, scalable, and practical solution for assessing numerical variability in deep learning, enabling researchers and practitioners to quantify and manage floating-point uncertainty without compromising performance or computational efficiency.
\end{abstract}


\section{Introduction}

Many scientific domains have increasingly adopted
deep learning (DL) models for computational analysis. However, these models
often rely on numerical computations that are sensitive to variations in
floating-point precision and rounding modes. Understanding numerical
variability in DL models is therefore essential for ensuring reliable and
reproducible outcomes. This is particularly critical in high-stakes
applications such as medical
imaging~\citep{des2023reproducibility,pepe2023numerical,vila2024impact}, remote
sensing~\citep{vicuna2021reducing}, and scientific
simulations~\citep{chatelain2022pytracer}.
Meanwhile, the importance of numerical variability is beginning to gain industry-wide
recognition. For example, variability analysis is now supported in Amazon Web
Services' (AWS) Neuron SDK~\citep{awsneuronrounding}, AWS Trainium chips, AMD
MI300 GPUs, Tesla D1s and Blackwell architecture NVIDIA
GPUs~\citep{elarar2025ms2a}. This reflects a growing demand for tools and
frameworks that enable low-level control over numerical behavior in AI systems.


Although exact methods such as interval arithmetic~\citep{hickey2001interval},
symbolic execution~\citep{solovyev2018rigorous}, and formal
verification~\citep{boldo2011flocq} exist to evaluate numerical accuracy, these
approaches often require extensive modifications to the codebase and do not
scale efficiently to complex deep learning workloads. Instead, we adopt
stochastic arithmetic, a technique that introduces controlled random
perturbations to floating-point operations. This enables statistical estimation
of the numerical variability without modifying the underlying model
architecture, making it more practical for large-scale applications.
Stochastic arithmetic encompasses various techniques, including Monte Carlo Arithmetic
(MCA)~\citep{parker1997monte}, CESTAC~\citep{brunet1986cestac} and Stochastic
Rounding~\citep{forsythe1959reprint}. To
leverage stochastic arithmetic, researchers have developed tools such as
Verificarlo~\citep{denis2016verificarlo}, Verrou~\citep{fevotte2016verrou}, the stochastic rounding library by Fasi and
Mikaitis~\citep{fasi2021algorithms} and
CADNA~\citep{jezequel2008cadna}. 

While stochastic arithmetic has been widely explored in numerical
analysis, it remains underused in deep learning, despite recent studies
demonstrating its potential to assess uncertainty in neural network training and
inference~\citep{faraone2019monte,kloberdanz2022deepstability,beuzeville2024deterministic,arar2025mixed}. 
A major practical challenge of stochastic arithmetic is the need to run programs multiple times, typically 10 times or more, to obtain stable statistical estimates of their variability. However, currently available stochastic arithmetic tools, especially Verrou, one of the more accessible stochastic arithmetic frameworks for DL, introduce slowdowns of 10× to 1000× on DL models with only a few million parameters. In such settings, collecting enough samples takes days or weeks, even with parallelization, and scaling these analyses to modern large language models becomes effectively infeasible. For numerical variability research to keep pace with rapidly growing DL architectures, it is therefore essential to develop tools that introduce minimal computational overhead. This motivates the design goals of Fuzzy PyTorch, which prioritizes speed and scalability so that numerical variability studies remain tractable even on increasingly large models.

Fuzzy PyTorch integrates stochastic arithmetic, more specifically Monte Carlo Arithmetic (MCA), into the PyTorch framework through a novel library
named \textbf{P}robabilistic
\textbf{R}ounding with \textbf{I}nstruction \textbf{S}et \textbf{M}anagement
(PRISM). PRISM
implements two modes: stochastic rounding (SR) and Up-Down rounding (UD). SR
bypasses exact operations and therefore preserves exact values by probabilistically rounding values based on their
proximity to representable floating-point numbers. Meanwhile, UD is a newly proposed rounding mode that is faster at the level of individual operations, as it randomly rounds up or down with equal probability. Both modes
are optimized using vectorized CPU instructions through the Highway
library~\citep{highway}, minimizing computational overhead.
Fuzzy PyTorch seamlessly integrates with PyTorch by extending the Verificarlo
compiler, providing a fast, practical framework for numerical variability analysis in deep learning. By integrating directly with the PyTorch execution pipeline and avoiding the major bottlenecks present in traditional approaches, such as the strict serialization in Verrou or the lack of vectorized support in CADNA and standard MCA, Fuzzy PyTorch enables full-scale analyses that were previously computationally prohibitive.
Compared to state-of-the-art tools, Fuzzy PyTorch achieves similar numerical-error characteristics but with markedly lower runtime, allowing researchers to evaluate numerical variability across full DL workflows. This efficiency makes large-scale, systematic studies of floating-point behavior feasible, directly supporting research on reproducibility, robustness, and numerical uncertainty in modern neural networks and helping advance reproducible and principled DL research.

This work proposes three main contributions:
\begin{enumerate}
  \item \textbf{PRISM:} We introduce PRISM, which implements fast probabilistic
        rounding methods for the systematic analysis of floating-point errors.
  \item \textbf{Stochastic arithmetic in PyTorch:} We seamlessly integrate stochastic arithmetic
        into PyTorch, enabling efficient and transparent instrumentation of deep
        learning operations.
  \item \textbf{Comparative evaluation with Verrou:} We perform a comprehensive
        evaluation against Verrou, a state-of-the-art tool for numerical variability
        analysis, showcasing the enhanced performance and flexibility of Fuzzy
        PyTorch on use cases ranging from digit classification with MNIST~\citep{lecun1998mnist}, whole brain MRI segmentation with the FastSurfer neuroimaging model~\citep{henschel2020fastsurfer}, and Parkinson's classification from speech data with the WavLM model~\citep{chen2022wavlm}.

\end{enumerate}

The remainder of this paper is structured as follows: Section~\ref{sec:implementation} details the design and implementation of Fuzzy Pytorch, including the UD rounding mode, the PRISM library implementation and PyTorch instrumentation.
Section~\ref{sec:results} presents results validating numerical accuracy and demonstrating 5–60× runtime speedups over Verrou. Section~\ref{sec:conclusion} concludes with discussion of limitations and future directions. Supplementary material, including additional information on existing rounding modes, use cases, the algorithm for probabilistic rounding and further statistical analysis of the harmonic series, is provided in the Appendix.

\section{Fuzzy Pytorch Design and Implementation}
\label{sec:implementation}

Fuzzy Pytorch implements a new rounding mode, Up-Down rounding (subsection~\ref{subsec:ud}), a faster alternative to stochastic rounding implemented in the PRISM library (subsection~\ref{subsec:prism}). 
The PRISM library implements the SR and UD rounding
modes and the modifications to
the Verificarlo compiler for compiling PyTorch with the PRISM library.

\subsection{Up-Down Rounding\label{subsec:ud}}

The Up-Down Rounding (UD) technique rounds the result of an already rounded floating-point operation to the next or previous floating-point number
with equal probabilities. Although UD rounding does not preserve exact
floating-point operations, it produces results similar to SR rounding on large
code bases, while generally being significantly faster. UD rounding is defined as:
\begin{align}
  \label{eq:ud}
  \circ_{\textsc{ud}}(x) = \begin{cases}
                             \circ_{\textsc{rn}}(x) - \epsilon(x) \, & \text{with probability } \frac{1}{2} \\
                             \circ_{\textsc{rn}}(x) + \epsilon(x)    & \text{with probability } \frac{1}{2} \\
                           \end{cases}
\end{align}

where $\epsilon(x)$ is the \emph{unit in the last
  place}~\citep{muller2018handbook} if $x \neq 0$ and 0 otherwise, and
$\circ_{\textsc{rn}}(x)$ is the IEEE-754 round-to-nearest with ties-to-even
rounding mode. Additional details and formal definitions of stochastic rounding, as well as the other rounding modes used for comparison to UD (CESTAC, and IEEE-754), are provided in Appendix~\ref{sec:num_var}.

\subsection{PRISM\label{subsec:prism}}

PRISM is a \texttt{C++} library that implements the SR
(Appendix~\ref{subsec:sr}) and UD rounding (sub-section~\ref{subsec:ud}) modes. We do not currently plan to support CESTAC, as it is not as commonly used as SR, but we consider it as a future work direction. PRISM
leverages the Highway library~\citep{google_highway} to use vectorized instructions
available on modern CPUs, thereby minimizing the overhead introduced by
stochastic arithmetic. Highway selects the best architecture target to generate
the most efficient code for the CPU, either at compile time (static dispatch)
or runtime (dynamic dispatch).

PRISM provides probabilistic rounding (SR and UD) for the floating-point
operations \(\{+,-,\div,\times,\surd,\text{Fused Multiply-Add (FMA)}\}\). The SR operators (except FMA) are implemented using the
rounding-mode-free algorithms by~\citep{fasi2021algorithms}.
We extend these algorithms to support the FMA instruction (described in Algorithm~\ref{algo:fma}). Our FMA implementation is inspired by Verrou and is
based on the ErrFmaNearest Algorithm by~\citep{boldo2010exact}.

\begin{algorithm}
  \caption{FMA With Stochastic Rounding Without the Change of the Rounding Mode}
  \begin{algorithmic}[1]
    \Function{FMA2}{$a \in \mathcal{F}$, $b \in \mathcal{F}$, $c \in
        \mathcal{F}$} \State Compute \(\varrho  = \circ_{\text{SR}}(a \cdot b + c)
    \in \mathcal{F}\) \State \(Z \gets \text{rand}()\) \State \(\sigma  \gets
    \circ_{\text{RN}}(a \cdot b + c)\) \State \((u_1, u_2) \gets
    \text{TwoProdFMA}(a, b)\) \State \((\alpha_1, \alpha_2) \gets
    \text{TwoSum}(c, u_2)\) \State \((\beta_1, \beta_2) \gets \text{TwoSum}(u_1,
    \alpha_1)\) \State \(\gamma \gets \circ_{\text{RN}}(
    \circ_{\text{RN}}(\beta_1 - r_1) - \beta_2 )\) \State \(\tau  \gets
    \circ_{\text{RN}}( \gamma + \alpha_2 )\) \State round \(\gets
    \text{SRround}(\sigma , \tau , Z)\) \State \(\varrho  \gets
    \circ_{\text{RN}}(r_1 + \text{round})\) \State \textbf{Return} result
    \EndFunction
  \end{algorithmic}
  \label{algo:fma}
\end{algorithm}

PRISM's interface offers functions for scalar and vector instructions,
supporting both static and dynamic dispatch. The static dispatch versions
accept vector types as inputs, while the dynamic dispatch versions accept
pointers to scalar types. Dynamic dispatch is necessary because vector types
may not be available at runtime (e.g., 512-bit AVX-512 registers on AVX2
architecture with 256-bit registers). Although slightly slower, dynamic
dispatch enhances portability, enabling x86-64 binaries to run on any
architecture.

Finally, PRISM supports multithreaded execution by assigning a separate random
generator to each thread, ensuring that concurrent executions do not share the
same seed state. This enables optimal performance without requiring any
synchronization mechanism and prevents correlations in the generated
floating-point perturbations across threads.


We modified the Verificarlo compiler~\citep{denis2016verificarlo} to use the PRISM library. 
Verificarlo replaces floating-point operations with generic calls to configurable backends
(e.g., MCA, IEEE, VPREC~\citep{chatelain2019automatic}) at the
LLVM~\citep{lattner2004llvm} Intermediate Representation (IR) level. In its
current version, Verificarlo serializes the vectorized instructions, which can
cause additional slowdowns. It adheres to the Interflop~\citep{defour2021interflop} interface, which exposes scalar arithmetic operations but not vectorized ones.
Specifically, we implemented new LLVM
instrumentation passes to replace scalar and vectorized floating-point
operations with calls to the PRISM library. We also ensured ABI compatibility
between the PRISM library and the source code to prevent incorrect register use
during argument passing.

\subsection{PyTorch Instrumentation\label{subsec:pytorch_instrumentation}}

We instrumented PyTorch version 2.2.1 with Verificarlo 2.0.0, using the PRISM
library as the backend. Verificarlo was built with LLVM version 7.0.0,
including support for FORTRAN code through LLVM's flang compiler. We used
Python 3.8.5. To ensure compatibility with Verificarlo, we modified only one
line in the PyTorch codebase. Specifically, we removed the \texttt{noexcept}
keyword from the move constructor of the \texttt{Module} class in
\texttt{torch/csrc/jit/api/module.h}. This adjustment was necessary to prevent
compilation errors related to LLVM compatibility but should no longer be
required with more recent LLVM versions.

To achieve complete instrumentation, we compiled the open-source BLAS and
LAPACK libraries~\citep{anderson1999lapack} with Verificarlo, replacing the
proprietary Intel MKL library as the default BLAS implementation.
Architecture-specific instructions (\texttt{-march=native}) were enabled to
leverage vectorized operations. Additionally, the ONNX~\citep{onnx} runtime was
compiled with the PRISM library to ensure comprehensive instrumentation of the
entire model execution. We disabled the use of the Intel MKL DNN~\citep{onednn}
library to avoid reliance on proprietary software. We did not instrument the
protobuf~\citep{protobuf} third-party library to avoid perturbing model serialization. We conducted the experiments using the software versions listed above. While we have also instrumented PyTorch 2.6.0 with Python 3.10 and LLVM 11.0.0 via Verificarlo 2.2.0 (the code is available for compilation in our documentation), these were not used in the experiments reported in this paper.


We excluded specific functions from instrumentation because they were
susceptible to producing erroneous outputs under our rounding modes. Correct
instrumentation alternatives are under development. In particular, PyTorch’s
exponential and logarithmic functions in the SLEEF (SIMD Library for Evaluating
Elementary Functions) third-party library exhibited large output deviations when
perturbed, especially when input values approached zero. Similarly, the
\texttt{torch.argmax} operation became unreliable under UD rounding, likely
because approximate rounding can alter comparisons when input values are close,
affecting the selected index. As these functions are sensitive and integral to
correct model behavior, they were not instrumented to ensure the reliability of
the current results.

\subsection{Availability of Data and Code}

The code and data used in this study are available on GitHub at 
\url{https://github.com/big-data-lab-team/fuzzy-pytorch}. 
The repository
includes the PRISM library, the modified Verificarlo compiler, the Dockerfile
to build Fuzzy PyTorch, and the scripts used to run the experiments.

\subsection{Computational Infrastructure}

All experiments except for the WavLM experiment were conducted on a server
equipped with 8 compute nodes with 32 cores Intel(R) Xeon(R) Gold 6130 CPU @
2.10GHz 22MB cache L3. The WavLM experiment was conducted on 
the Narval cluster
from \'Ecole de Technologie Sup\'erieure (ETS, Montr\'eal), managed by Calcul
Qu\'ebec and The Digital Alliance of Canada 
which includes AMD Rome 7502, AMD
Rome 7532, and AMD Milan 7413 CPUs with 48 to 64 physical cores, 249~GB to
4000~GB of RAM and Linux kernel 3.10.



\section{Results}
\label{sec:results}

We evaluated the
accuracy, runtime efficiency, and numerical variability of Fuzzy PyTorch using Verrou, CADNA, and the stochastic rounding library of Fasi and Mikaitis as baselines. We performed (1) a sanity check with the harmonic series, a classical example in numerical analysis, (2) a computational
overhead assessment with the NAS Parallel Benchmarks, demonstrating Fuzzy PyTorch’s
performance in an HPC context, and (3) runtime performance and
numerical variability assessments in three practical deep-learning applications: MNIST
digit classification, FastSurfer CNN brain segmentation, and WavLM-based speech classification.

\subsection{Numerical Error Estimation Baselines}
\label{subsec:num_tools}

We compared the PRISM library with several state-of-the-art tools for numerical
variability analysis.
Verrou~\citep{nethercote2007valgrind} is a Valgrind-based Dynamic Binary Instrumentation tool that perturbs floating-point operations by replacing them with custom rounding functions, supporting modes such as Stochastic Rounding (SR, called average rounding) and asynchronous CESTAC (called random rounding), and can instrument binaries without recompilation, though multithreading is serialized due to Valgrind. The Verificarlo MCA backend~\citep{parker1997monte} implements Monte Carlo Arithmetic by injecting uniform perturbations with a user-defined virtual precision~$t$, performing computations at twice the target precision to avoid double rounding, and offering modes including Random Rounding (equivalent\footnote{The MCA RR mode
  is biased for inputs close to a power of two, see~\citep{de2022high}.} to SR when $t=p$), but incurs overhead from serializing vectorized instructions and using 128-bit arithmetic for binary64. CADNA~\citep{jezequel2008cadna} provides a CESTAC-based synchronous stochastic arithmetic library via overloaded stochastic types that propagate three perturbed values per operation and estimate significance loss (e.g., ``computational zero''), with OpenMP support for threads but no vectorized instruction support. Finally, the stochastic rounding library of Fasi and Mikaitis (FM)~\citep{fasi2021algorithms} offers C functions implementing SR for scalar floating-point operations (excluding FMA), requiring explicit replacement of each arithmetic operation and lacking vectorized interfaces. We
will refer to this library as FM SR in the remainder of the paper.

\subsection{Harmonic Series Validates Expected Variability Patterns}

To evaluate the accuracy of the UD and SR rounding modes, we analyzed the
harmonic series \(\sum_{i=1}^n \frac{1}{i}\) for \(n\) ranging from \(10^2\) to
\(10^7\). While this series is divergent in real arithmetic, it converges in
floating-point arithmetic for \(n \geq 2^{48}\) within IEEE-754 binary32 format due to numerical absorption~\citep{malone2013does}. We performed computations in IEEE-754 binary32 format and compared PRISM against established baselines: CESTAC (via CADNA), MCA RR (via Verificarlo), Verrou (SR and CESTAC modes) and FM SR. A binary64 computation served as the reference value.

\begin{figure}[h]
  \includegraphics[width=\textwidth]{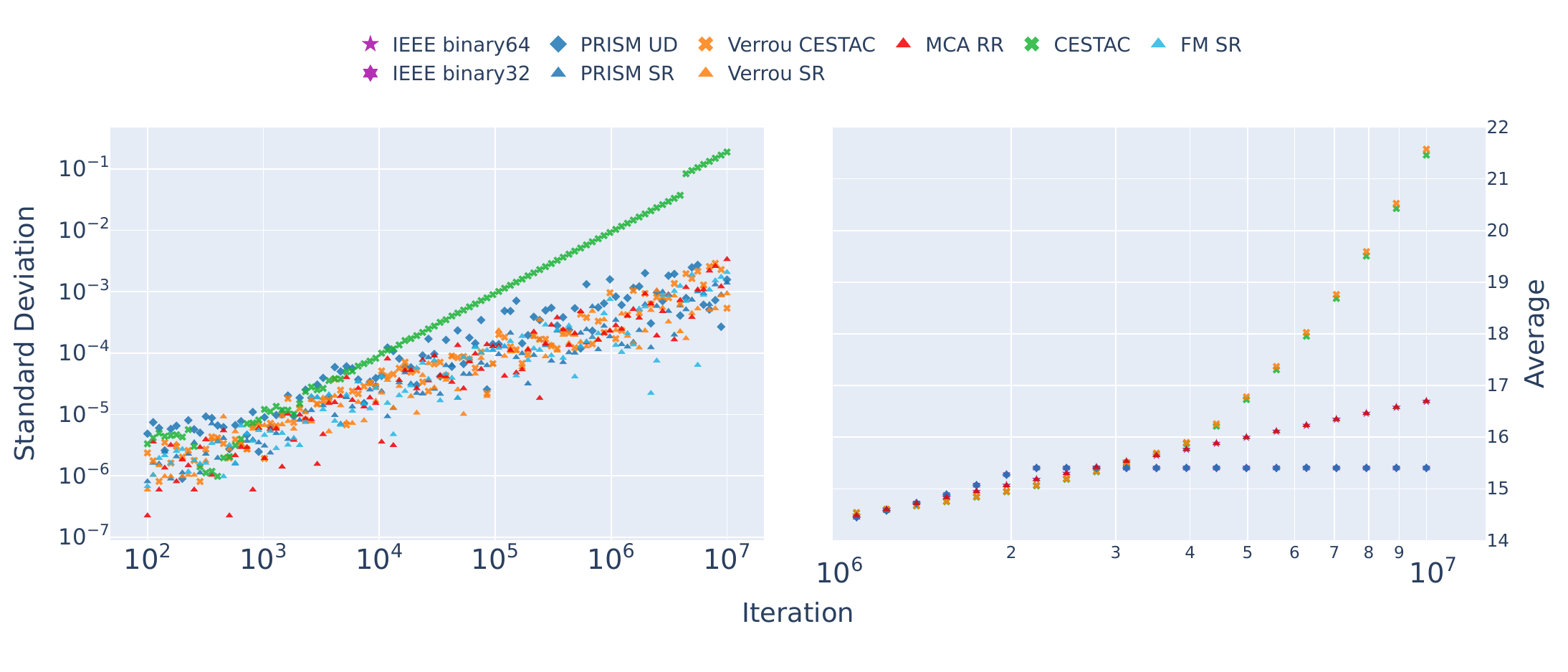}
  \caption{Comparison of probabilistic rounding on the harmonic series example. }
  \label{fig:harmonic_series}
\end{figure}

Figure~\ref{fig:harmonic_series} presents
the standard deviation and mean values obtained from three repetitions
per mode (the three internal values extracted from a single run for CADNA). The 
results show that PRISM SR 
exhibits variability comparable to MCA RR, Verrou SR, and FM SR (Levene's test,
$p=0.29$; see Extended Data Table~\ref{tab:harmonic_stats}). In contrast, UD rounding displays higher variability, which is expected as it applies perturbations to values that have already been rounded. Finally, CADNA
exhibits the highest variability as it ensures different rounding directions across its three internal repetitions.

As shown in Figure~\ref{fig:harmonic_series}, the mean values exhibit three
distinct behaviors:
\begin{enumerate}
  \item The MCA SR, Verrou SR, PRISM SR, and FM SR modes closely approximate the
        binary64 reference value. This aligns with expectations, as SR rounding
        converges to the expected value.
  \item The CESTAC and Verrou CESTAC modes diverge more rapidly. This behavior is
        consistent with the known bias introduced by CESTAC rounding.
  \item The PRISM UD mode converges toward the binary32 result, as expected,
        since UD rounding applies random perturbations to values that have already
        been rounded using the round-to-nearest mode.
\end{enumerate}

\subsection{PRISM UD Minimizes Runtime Slowdowns for NAS Parallel Benchmarks}

We evaluated the runtime overhead introduced by PRISM using the \texttt{C++} NAS
Parallel Benchmarks version 3.4.1~\citep{loff2021parallel}  (Appendix Table~\ref{tab:npb}).
The Integer Sort (IS) benchmark was excluded, as it does not involve
floating-point operations.

Experiments were performed using the serial implementation on dataset classes
\textit{S} and \textit{A}, corresponding to the smallest benchmark workloads.
PRISM SR and UD modes were compared
against CESTAC (via CADNA), MCA RR (via Verificarlo), Verrou (SR and CESTAC modes) and FM SR.
All benchmarks were compiled with \texttt{-march=haswell -maes} to enable AVX2
instructions and executed using IEEE-754 binary64 arithmetic.
Runtimes were averaged over three independent repetitions.
An uninstrumented execution was used as the baseline.

\begin{figure}[htbp]
  \centering
  \includegraphics[width=\textwidth]{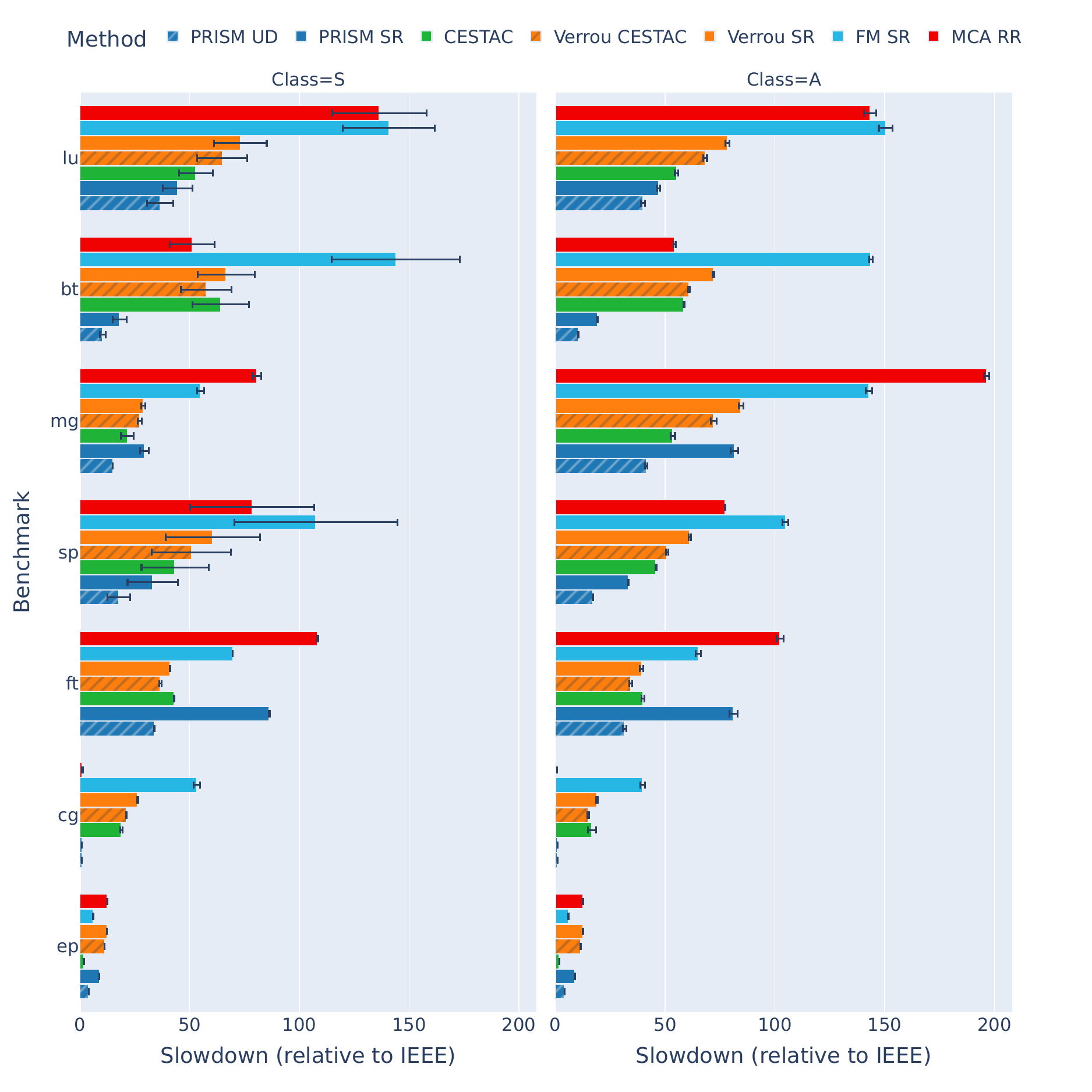}
  \caption{ Comparison of slowdowns across numerical variability analysis tools
    for NAS Parallel Benchmarks on dataset S and A, relative to the IEEE
    binary64 baseline.}
  \label{fig:npb_runtime}
\end{figure}

Figure~\ref{fig:npb_runtime} shows that PRISM SR induces runtime slowdowns that
are comparable to, or lower than, those observed for Verrou SR and CADNA.
PRISM UD consistently yields the lowest overhead among the evaluated tools.
These trends are consistent across dataset classes.

\subsection{Fuzzy PyTorch Achieves Significant Runtime Speedup}

\begin{figure}[ht]
  \centering
  \includegraphics[width=.8\linewidth]{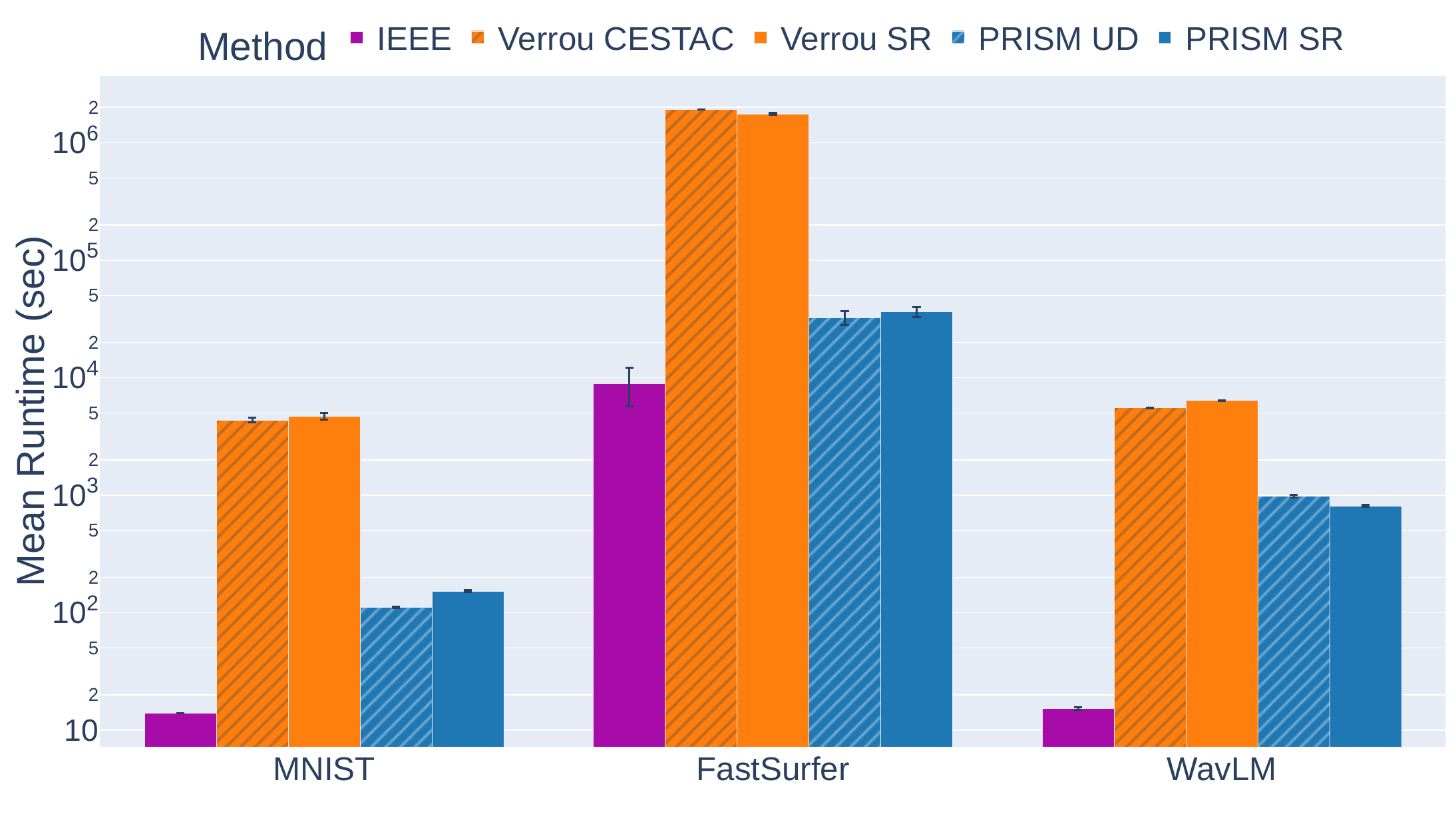}
  \caption{Comparison of instrumentation runtimes across DL models}
  \label{fig:runtime}
\end{figure}

To further evaluate Fuzzy PyTorch's efficiency in practical DL workflows, we measured inference runtime for MNIST, FastSurfer and WavLM  (see model descriptions in Appendix~\ref{sec:use_cases}). We compared PRISM UD and SR modes against Verrou's CESTAC and SR modes. All experiments
were executed using a single thread, as Verrou enforces serialization of multithreaded
execution.

As seen in Figure~\ref{fig:runtime}, Fuzzy PyTorch achieves speedups ranging
from $5\times$ to $60\times$ for UD mode and $7\times$ to $49\times$ for SR mode
compared to Verrou. This improvement is notable 
considering that Verrou's PyTorch instrumentation relies on the highly
optimized Intel MKL library, whereas Fuzzy PyTorch uses the standard Netlib BLAS/LAPACK implementation.

In contrast to the NAS Parallel Benchmarks, where PRISM UD consistently
outperformed PRISM SR, the relative performance of UD and SR varies across deep
learning workloads.
For WavLM, SR achieves a larger speedup ($7.89\times$ relative to Verrou) than UD ($5.65\times$). 
UD retains an advantage for the CNN-based architectures FastSurfer (UD: $60.43\times$, SR: $49.07\times$) and MNIST (UD: $39.22\times$, SR: $30.81\times$).

We attribute the UD slowdown on WavLM to its transformer-based architecture.
Transformer models are computationally more complex than CNNs: 
self-attention layers introduce quadratic cost in sequence length with
heavier intermediate memory usage. 
Unlike UD which incurs a constant perturbation cost for every operation, SR 
bypasses exact operations, reducing its relative overhead.

\subsection{Comparable Numerical Variability between Fuzzy PyTorch and Verrou}


\begin{figure}[htbp]
  \centering
  \begin{subfigure}[t]{0.48\textwidth}
    \centering
    \includegraphics[width=\textwidth]{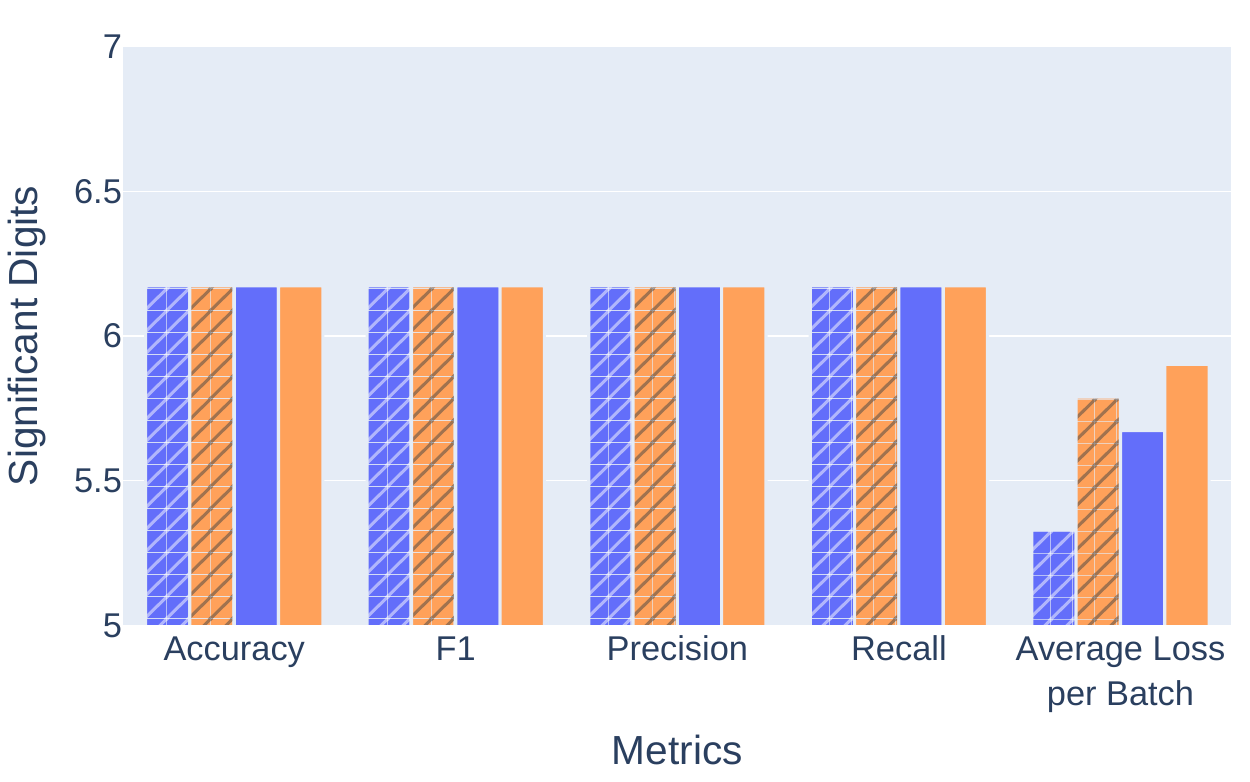}
    \caption{MNIST}
    \label{fig:mnist_sig_metrics}
  \end{subfigure}
  \hfill
  \begin{subfigure}[t]{0.48\textwidth}
    \centering
    \includegraphics[width=\textwidth]{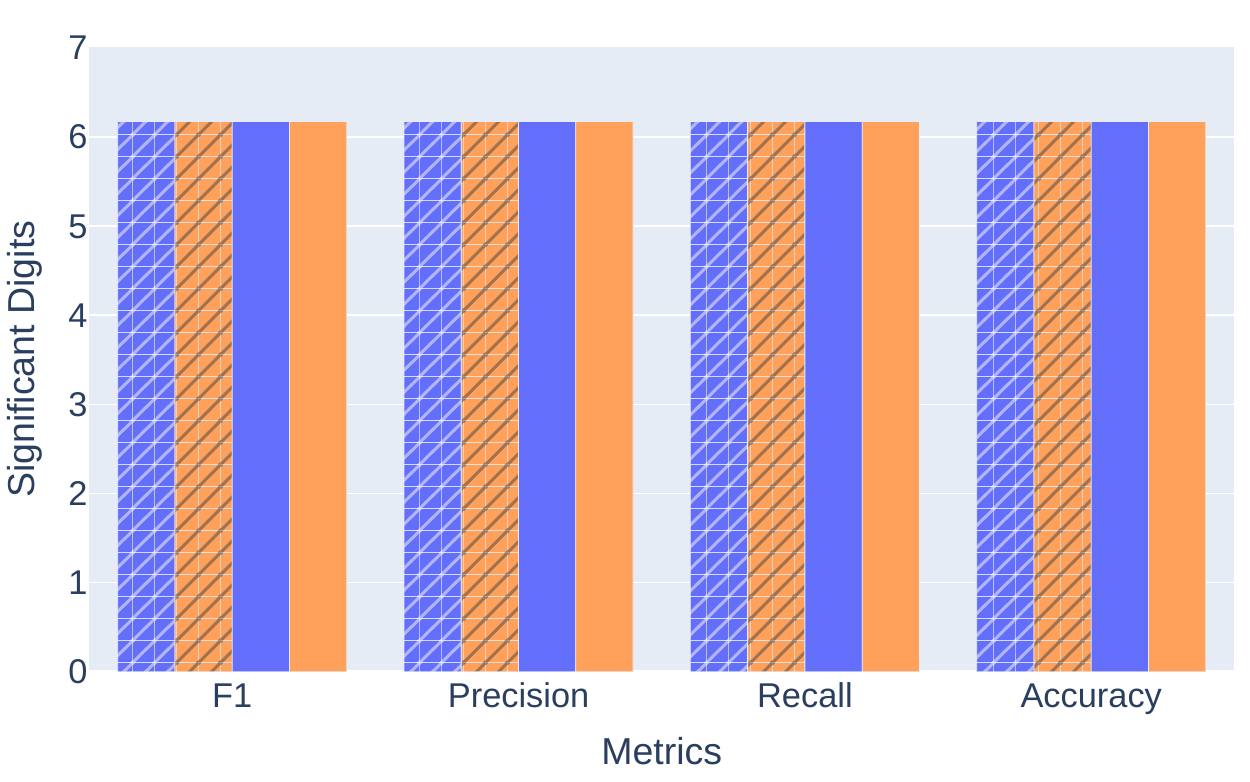}
    \caption{WavLM}
    \label{fig:wavlm_sig_metrics}
  \end{subfigure}

  \vspace{0.8em} 

  \begin{subfigure}[t]{0.6\textwidth}
    \centering
    \includegraphics[width=\textwidth]{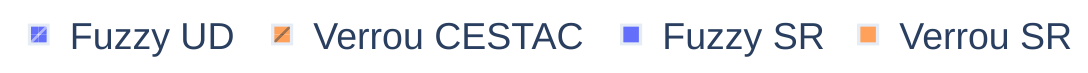}
    \label{fig:sig_metrics_legend}
  \end{subfigure}

  \caption{Significant digits across MNIST and WavLM model metrics for different instrumentation tools.}
  \label{fig:mnist_wavlm_comparison}
\end{figure}

To assess Fuzzy PyTorch beyond runtime performance, we evaluated the numerical variability
it introduces compared to Verrou across MNIST, FastSurfer and WavLM.

For MNIST in Figure~\ref{fig:mnist_sig_metrics}, we evaluated standard classification metrics, including accuracy and weighted precision, recall, and F1 score, and quantified numerical variability using the significant digits metric~\citep{sohier2021confidence}. In binary32 arithmetic, the theoretical upper bound is 7.23 significant digits; across all metrics we observed a maximum of approximately 6.17 significant digits, with the loss function exhibiting noticeably higher variability. Variability is effectively confined to the loss across both Fuzzy PyTorch and Verrou, which is expected given that MNIST is a well-solved task with highly stable predictions. UD rounding consistently introduces greater variability than stochastic rounding for both tools. Fuzzy PyTorch shows slightly lower significant digits overall, which we attribute to its instrumentation of AVX-512 vector instructions, allowing a wider class of floating-point operations to be perturbed compared to Verrou.

In the WavLM use case (Figure~\ref{fig:wavlm_sig_metrics}), as with MNIST, we evaluated numerical variability across
accuracy and the weighted variants of precision, recall, and F1 score. On
average, we observe 6.17 significant digits across all performance metrics,
indicating high numerical stability—comparable to that observed across IEEE
executions.

\begin{figure}[htbp]
  \begin{subfigure}[b]{\linewidth}
    \centering
    \begin{tabular}{c}
      \includegraphics[width=0.8\linewidth]{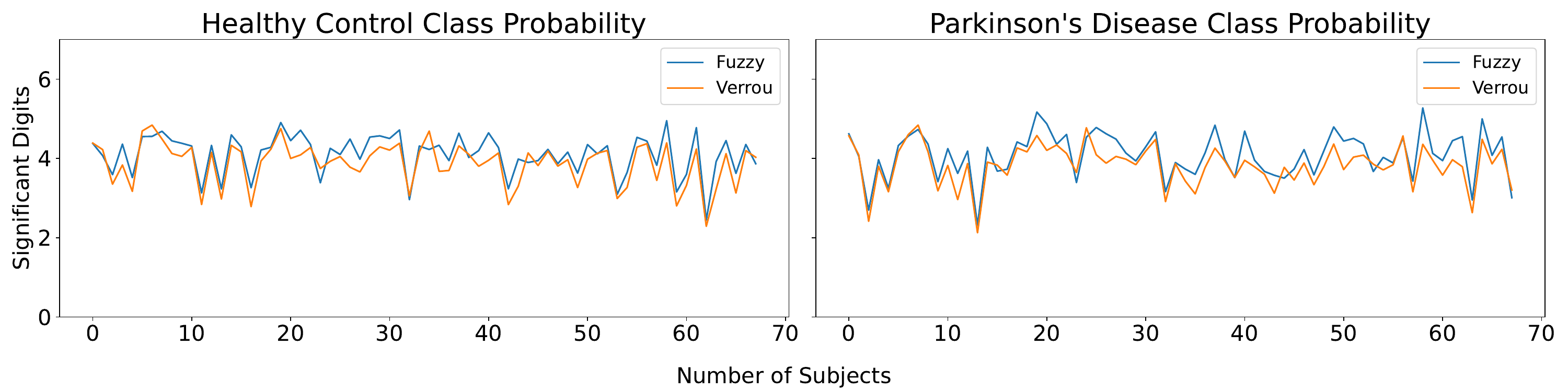} \\
      \scriptsize{Stochastic rounding mode}
      \\
      \includegraphics[width=0.8\linewidth]{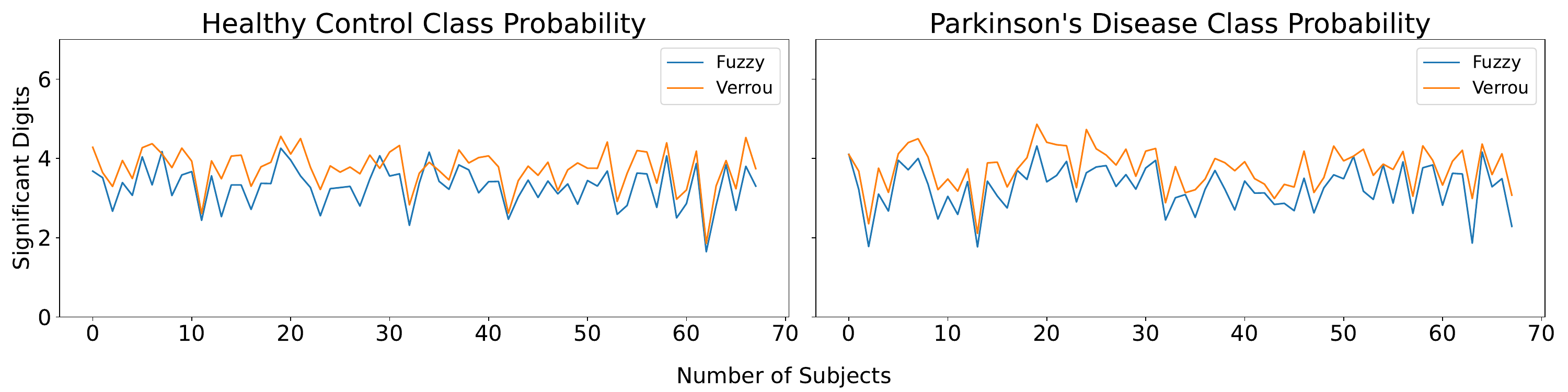}
      \\
      \scriptsize{Up-Down rounding mode}
    \end{tabular}

  \end{subfigure}
  \caption{Significant digits for rounding modes across WavLM model's class probabilities}
  \label{fig:wavlm_class_prob}

\end{figure}

To investigate whether this apparent stability conceals underlying instability,
we analyzed the model’s output probabilities before the final max operation. As
shown in Figure~\ref{fig:wavlm_class_prob}, the number of significant digits
drops, averaging around 4 across all modes and tools, with a standard deviation
of approximately half a digit. This suggests that some numerical instability is
indeed present but is masked by the final max operation. Consistent with
findings from previous use cases, we also note that PRISM UD mode
introduces the highest level of numerical perturbation—even surpassing Verrou’s CESTAC mode.

\begin{figure}[ht]
  \centerline{\includegraphics[width=\textwidth]{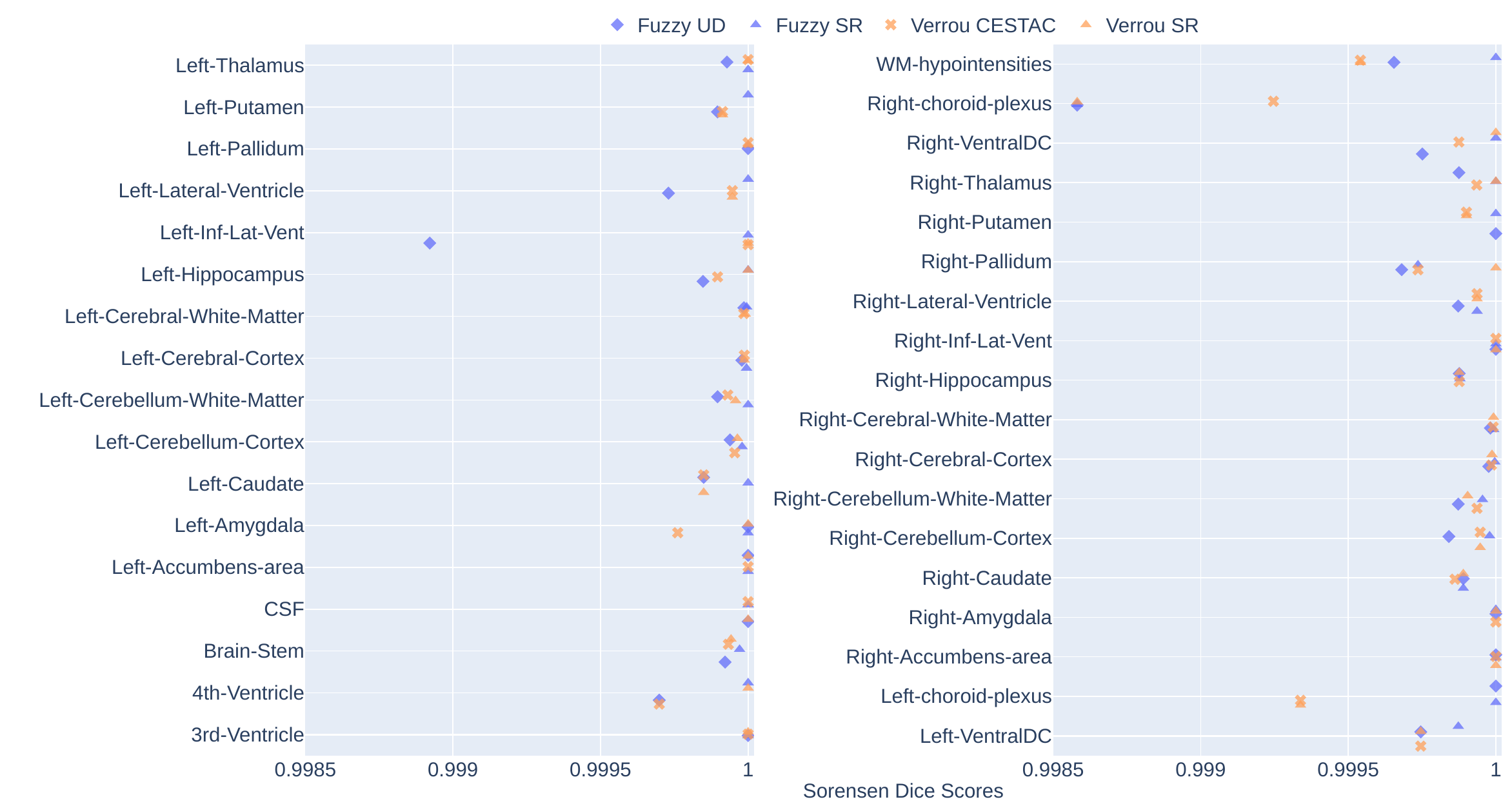}}
  \caption{ Minimum Sørensen-Dice score across instrumentation tools and
    different labelled brain regions.
    \label{fig:fastsurfer_dice}}
\end{figure}

For the FastSurfer use case, we assess variability at inference using the
minimum Sørensen–Dice scores between MCA iterations
(Figure~\ref{fig:fastsurfer_dice}). The minimum Sørensen–Dice score captures the
most extreme cases of variability across brain regions, offering a global
measure of segmentation consistency. Across all modes, coefficients remain
extremely high, with the lowest observed value approaching 0.9985. Variability magnitudes are similar across methods, though slightly
higher for PRISM UD, likely due to its lack of preservation of exact operations.
Similarly to the MNIST results, PRISM UD shows the highest variability but still
maintains comparable segmentation accuracy. These results confirm that our
MCA-based instrumentation for FastSurfer operates correctly, producing
consistent and interpretable variability measurements across modes.

For all use cases, we verified that no random processes were present after fixing the random seeds by running multiple iterations of the IEEE implementations of each model. In FastSurfer, this yielded a perfect minimum Sørensen–Dice score (1.0, standard deviation 0), confirming determinism. For MNIST and WavLM, all metrics, including loss, retained 6.17 significant digits.

\subsection{Model Embeddings Are Comparable}

\begin{figure}[htbp]
  \begin{subfigure}[b]{\linewidth}
    \centering
    \begin{tabular}{c}
      \includegraphics[width=\linewidth]{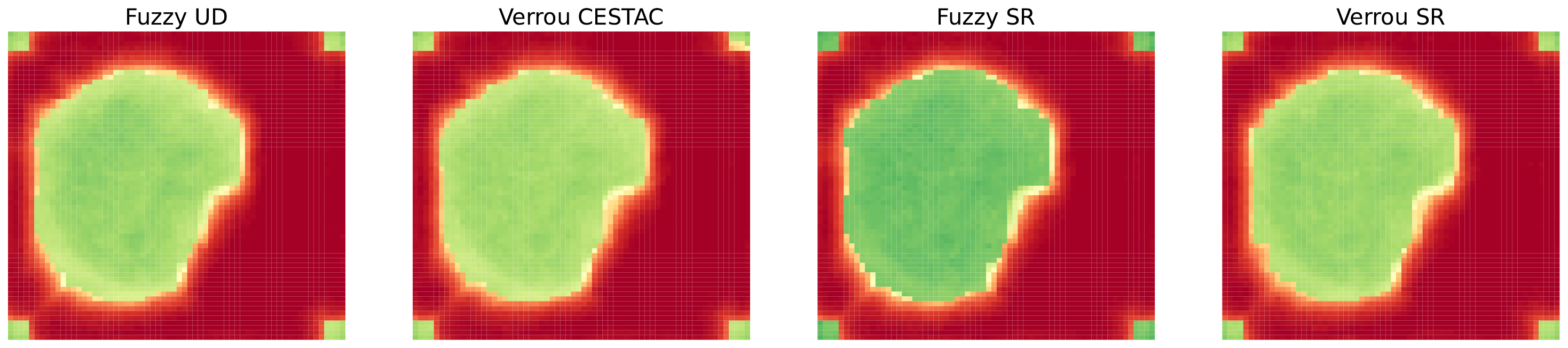} \\
      \scriptsize{FastSurfer}                                               \\
      \includegraphics[width=\linewidth]{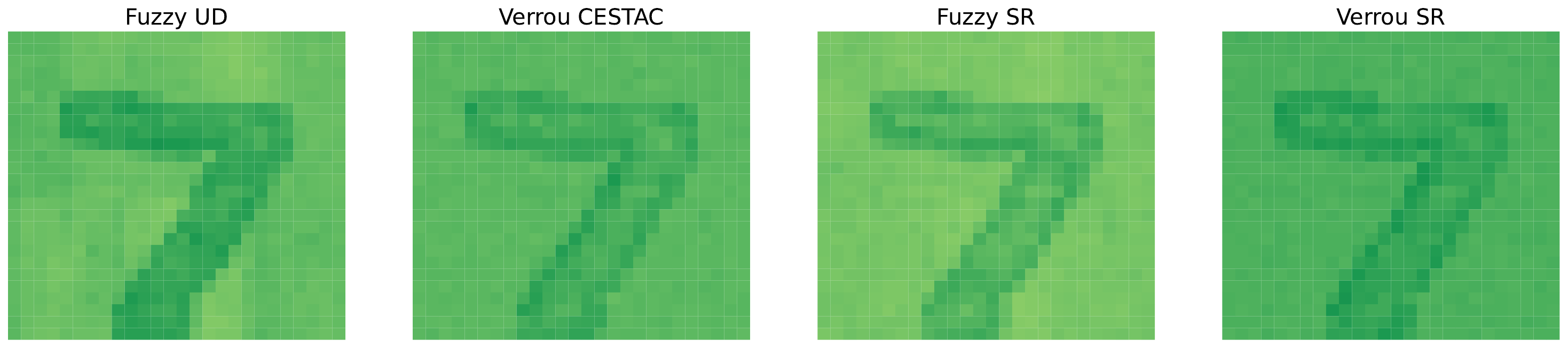}      \\
      \scriptsize{MNIST}                                                    \\
      \includegraphics[width=\linewidth]{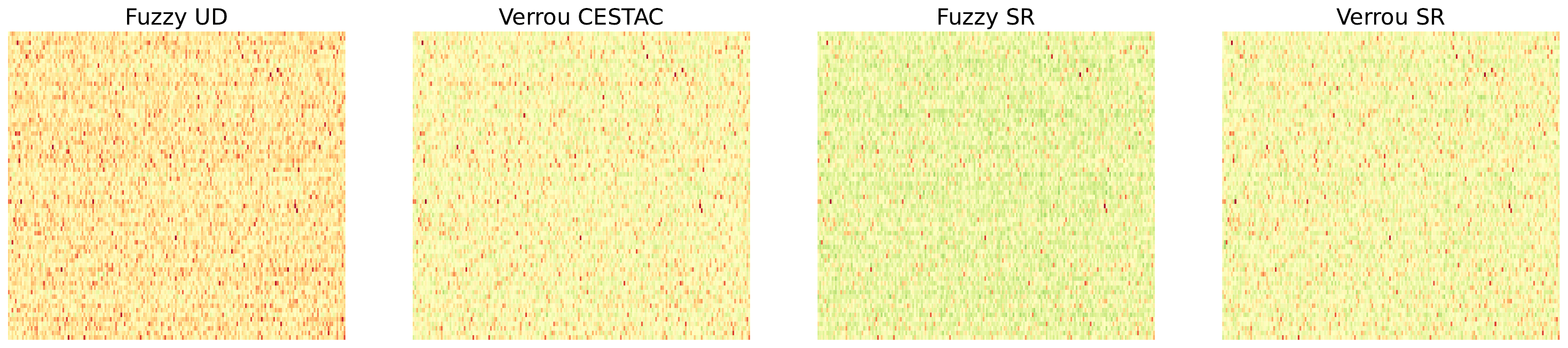}      \\
      \scriptsize{WavLM}
    \end{tabular}
    \centering
    \includegraphics[width=0.7\textwidth]{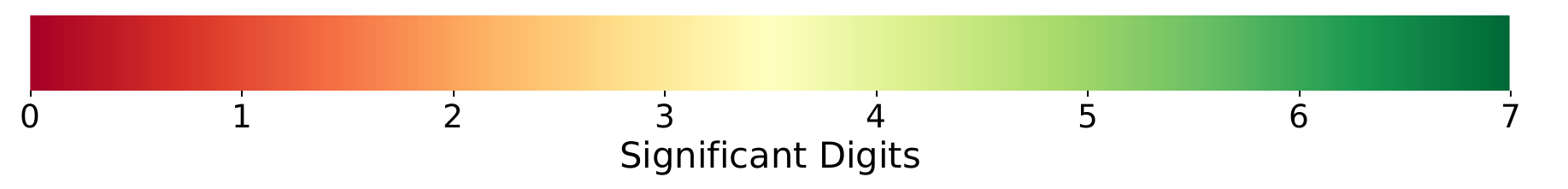}
  \end{subfigure}
  \caption{Model embeddings across stochastic arithmetic implementations }
  \label{fig:embeddings}

\end{figure}

To verify that numerical perturbations propagate correctly through the model architectures, we analyzed the variability of internal feature maps (embeddings). Figure~\ref{fig:embeddings} illustrates one embedding from each use case: the
output of the second decoder block for FastSurfer, the first convolutional
layer for MNIST and the output of the ECAPA-TDNN transformer component of the WavLM model.

Numerical variability followed consistent, task-dependent patterns across all tools and rounding modes, although, for WavLM, we can visibly see the slightly higher numerical uncertainty with Fuzzy PyTorch's UD mode compared to the other modes. 

While the numerical uncertainty within the
MNIST and WavLM embeddings aligns with the observed stability of its outputs,
FastSurfer presented a notable discrepancy: substantial uncertainty in the background of
its embeddings despite its stable outputs. Further investigation uncovered that this instability was confined to the
background region outside the brain and originated from unstable indices being
generated by the max-pooling operation across stochastic arithmetic iterations. Post-processing
steps in FastSurfer effectively masked the background, thereby mitigating the
impact of these instabilities on the final outputs for this specific use case.
This finding directly motivated the development of Conservative \& Aggressive NaNs,
two approaches for leveraging numerical uncertainty into computational
efficiency gains while preserving model performance~\citep{nanpool}.

Overall, these findings demonstrate consistency between the global patterns
observed in Fuzzy PyTorch and Verrou's results, reinforcing the reliability of
Fuzzy PyTorch in assessing variability throughout DL models.

\section{Conclusion}
\label{sec:conclusion}


Fuzzy PyTorch is a framework designed to evaluate numerical variability in DL
models, addressing the challenges posed by floating-point arithmetic
limitations. By leveraging vectorized CPU instructions via the Highway library,
it minimizes computational overhead while maintaining flexibility. The
framework supports two rounding modes, SR and UD, allowing researchers to
balance precision and computational efficiency, making it a versatile tool for
enhancing model robustness and reproducibility.
Across harmonic series tests and deep learning tasks, SR consistently provides the most accurate and stable results, while UD drifts toward lower precision and CESTAC diverges due to bias. Fuzzy PyTorch reproduces these variability patterns and achieves accuracy comparable to state-of-the-art tools such as Verrou across MNIST classification, FastSurfer segmentation, and WavLM Parkinson’s detection.

Fuzzy PyTorch crucially achieves substantial speedups, with minimal slowdowns for PRISM SR and especially PRISM UD in NAS parallel benchmarks. In deep learning tasks, it reaches up to $60\times$ acceleration—despite using less optimized CPU libraries (OpenBLAS and LAPACK versus Intel MKL). While PRISM UD is fastest in smaller benchmarks, its advantage decreases in DL experiments, likely because it does not skip exact operations. Importantly, such speed is not only beneficial for small numerical tests but becomes essential in large-scale DL workloads, where evaluating numerical variability can otherwise be prohibitively slow. The ability to run multiple stochastic executions at scale is critical for feasible, timely numerical analysis. This efficiency gain enables the scalable, systematic evaluation of numerical variability in large DL models, a capability unmatched by existing methods.

As numerical variability becomes increasingly recognized across the AI industry, the availability of system-level controls—such as AWS Neuron's hardware-supported rounding modes—highlights the pressing need for software frameworks like Fuzzy PyTorch. These frameworks make it possible to perform fine-grained evaluations of numerical behavior across diverse architectures, from CPUs to accelerators, and ensure that both academic and industrial workflows benefit from deeper guarantees of computational robustness.
By systematically analyzing variability, we can identify sources of instability that can be leveraged for practical improvements. For example, during our work with the FastSurfer CNN, Fuzzy PyTorch revealed regions of numerical instability that we subsequently exploited to implement Conservative \& Aggressive NaNs, yielding significant computational efficiency gains~\citep{nanpool}. This illustrates how efficient variability analysis not only ensures reproducibility but also enables optimization and the development of novel techniques that improve performance in large-scale deep learning models.

While our current implementation is CPU-based, the main conclusions from CPU
testing are expected to generalize to GPU architectures. Future work, involves extending
the framework to GPU architectures, although it poses significant technical challenges. The
GPU tooling ecosystem is far more fragmented than its CPU counterpart, requiring
specialized approaches for different vendor platforms. A successful GPU
implementation would need to address multiple compilation targets and runtime
environments. For NVIDIA GPUs, dynamic instrumentation frameworks like
NVBit~\citep{villa2019nvbit} could enable analysis of closed-source libraries,
but this approach is complex and platform-specific. Alternatively,
compiler-based solutions using IREE~\citep{liu2022tinyiree} with specific
compilation targets could offer broader hardware support, though this would
require substantial development effort to integrate with existing PyTorch
workflows. Nonetheless, we expect to make the Fuzzy PyTorch framework applicable to GPU-based deep learning workloads
once Verificarlo extends its support to GPU architectures.
Future directions for Fuzzy PyTorch, beyond GPU support, include exploring the impact of numerical
variability on diverse DL model architectures, optimizing SR mode performance,
extending PRISM with additional floating-point formats and specialized DL
instructions.

In summary, Fuzzy PyTorch provides an efficient, reliable, and versatile tool
for assessing numerical variability in DL models. It empowers researchers to
deepen their understanding of numerical behavior in DL models and enhances
their ability to develop robust and reproducible systems.

\bibliography{main}
\bibliographystyle{tmlr}

\newpage

\appendix

\section{Numerical Variability Estimation}
\label{sec:num_var}
Measuring numerical variability in DL models can involve a family of
techniques that introduce randomness into floating-point computations. These
methods rely on non-deterministic rounding. Unlike standard IEEE-754 rounding
modes, this approach rounds to either of the two closest floating-point numbers
based on computed probabilities. The specific stochastic arithmetic technique
is determined by how this probability is computed. 
We use the term Probabilistic Rounding (PR), introduced in~\citep{higham2019new}, as an umbrella term for this class of techniques. Moreover, because IEEE rounding is deterministic rather than probabilistic, we present it in its own subsection to emphasize this difference.


\subsection{IEEE-754 Rounding}
\label{subsec:ieee_rounding}

The IEEE-754 round-to-nearest mode, also known as round-to-nearest,
ties-to-even, is the default rounding mode used in most floating-point hardware
and software implementations. When a real number cannot be exactly represented
in floating point, it is rounded to the closest representable number. If the
number lies exactly halfway between two floating-point values, the tie is
broken by rounding to the one with an even least significant bit (i.e., the one
whose mantissa ends in 0).

Formally, the rounding function $\circ_{\textsc{rn}}(x)$ maps $x \in
  \mathbb{R}$ to the nearest floating-point number $f \in \mathcal{F}$ such that:
\[
  \circ_{\textsc{rn}}(x) = \arg\min_{f \in \mathcal{F}} |x - f|
\]
with ties resolved to the floating-point number $f$ such that the significand
of $f$ is even. This method minimizes rounding bias over repeated operations
and is the most widely adopted deterministic rounding strategy defined by the
IEEE-754 standard~\citep{ieee2008ieee}, which specifies several modes including round-to-nearest (tie-to-even or tie-to-odd), rounding toward ±$\infty$, and rounding toward zero.

\subsection{Probabilistic Rounding}
\label{subsec:prob_rounding}

Let $\mathcal{F}$ be the set of normalized binary floating-point numbers with
elements $x= (-1)^s . m . 2^e$, where $s \in \{0,1\}$ is the sign bit, $2^{p-1}
  \leq m < 2^p $ is the significand, $e \in \mathbb{Z}$ is the exponent and $p$
is the precision. Let the rounding functions $\ceil{x}: \mathbb{R} \to
  \mathcal{F}$ and $\floor{x}: \mathbb{R} \to \mathcal{F}$ be such that
$\floor{x} = \max \{y \in \mathcal{F} \, | \, y \leq x\}$ and $\ceil{x} = \min
  \{y \in \mathcal{F} \, | \, y \geq x\}$, which return the previous and next
representable floating point numbers to $x$. Probabilistic Rounding can be
defined as:
\begin{align}
  \circ_{\textsc{pr}}(x, p_{\circ}) = \begin{cases}
                                        \floor{x}\, & \text{with probability } p_{\circ}     \\
                                        \ceil{x}    & \text{with probability } 1 - p_{\circ} \\
                                      \end{cases}
\end{align}

where $p_{\circ}: \mathbb{R} \to [0,1]$.

The existing stochastic arithmetic techniques can then be reinterpreted with
the PR definition.

\subsubsection{CESTAC Rounding\label{subsec:cestac}}

The Contr\^{o}le et Estimation STochastique des Arrondis de Calculs (CESTAC)
technique simulates the roundoff error in floating-point arithmetic by rounding
upward or downward the result of each floating-point operation with equal
probabilities.
\begin{align}
  \label{eq:cestac}
  \circ_{\textsc{cestac}}(x) = \begin{cases}
                                 \floor{x}\, & \text{with probability } \frac{1}{2} \\
                                 \ceil{x}    & \text{with probability } \frac{1}{2} \\
                               \end{cases}
\end{align}

CESTAC is implemented in the CADNA
library~\citep{jezequel2008cadna}.\vspace{.25cm}

\subsubsection{Stochastic Rounding\label{subsec:sr}}

The Stochastic Rounding~\citep{forsythe1959reprint} (SR) technique rounds the
result of each floating-point with a probability that depends on the distance
between the exact value and the two closest representable floating-point
numbers. The probability is computed as:
\begin{align}
  \label{eq:sr}
  \circ_{\textsc{sr}}(x) = \begin{cases}
                             \floor{x}\, & \text{with probability } p_{\textsc{sr}}       \\
                             \ceil{x}    & \text{with probability }\: 1 - p_{\textsc{sr}} \\
                           \end{cases}
\end{align}
where $p_{\textsc{sr}}(x)$ is defined as:
\[
  p_{\textsc{sr}}(x) = 1 - \frac{x - \floor{x}}{\ceil{x} - \floor{x}}
\]

SR implementations include, but are not limited to, the Random Rounding (RR) mode
of MCA in Verificarlo, the
implementation of Fasi and Mikaitis and the average
rounding mode of Verrou.


\section{NAS Parallel Benchmarks description}

\begin{table}[H]
  \begin{center}

    \begin{tabular}[c]{|c|c|c|c|c|c|c|c|c|}
      \hline
      \textbf{Benchmark} & \textbf{Description}               \\
      \hline
      bt                 & Block Triangular Solver            \\
      cg                 & Conjugate Gradient                 \\
      ep                 & Embarrassingly Parallel            \\
      ft                 & Fast Fourier Transform             \\
      lu                 & Lower-Upper Symmetric Gauss-Seidel \\
      mg                 & Multi-Grid Solver                  \\
      sp                 & Scalar Pentadiagonal Solver        \\
      \hline
    \end{tabular}
    \caption{NAS Parallel Benchmarks description\label{tab:npb}}
  \end{center}
\end{table}

\section{Deep Learning Use Cases}
\label{sec:use_cases}

We evaluated the accuracy and performance of Fuzzy PyTorch across 
three deep-learning models
demonstrating real-world applicability —MNIST (handwritten digit
classification), WavLM (applied to speech-based Parkinson’s disease
identification), and FastSurfer (brain segmentation using CNN).

\paragraph{MNIST}
To evaluate the performance and behavior of Fuzzy PyTorch, we conducted
experiments on a small CNN trained on the MNIST dataset~\citep{lecun1998mnist}. The model architecture
consisted of convolutional, ReLU, max-pooling, convolutional, ReLU, dropout,
flatten, linear, ReLU, dropout, and linear layers, followed by a log-softmax
output layer. The task, a classification problem to identify digits from the
input images, is well-established and widely considered solved. This experiment
served as a baseline to demonstrate the feasibility and potential benefits of
Fuzzy PyTorch in a controlled, well-understood context.
In order to quantify the numerical variability in MNIST, we use the significant bit metric~\citep{parker1997monte,sohier2021confidence}, which calculates the amount of shared information among the perturbed results; the more significant bits, the greater the precision. We estimate the number of significant bits using the non-parametric method described in~\citep{sohier2021confidence}, which is implemented in the
\texttt{significant\_digits} package~\citep{significantdigits-url}.

\paragraph{WavLM}
The model for Parkinson's identification from speech is
a pipeline composed of the WavLM Large~\citep{chen2022wavlm} model in a frozen configuration to extract features from the audio recordings that are fed to the Emphasized Channel
Attention, Propagation, and Aggregation Time Delay Neural Network
(ECAPA-TDNN)~\citep{desplanques2020ecapa} and a linear layer classifier followed
by a max operation to obtain binary classification to determine whether a
subject has Parkinson's disease or is a healthy control subject. The model was trained on the mPower speech dataset~\citep{bot2016mpower}. We will refer to this pipeline as WavLM in this work. For this use case, we also use significant bits to quantify the model's numerical variability.

\paragraph{FastSurfer}

\begin{figure}[H]
  \centering
  \includegraphics[width=\linewidth]{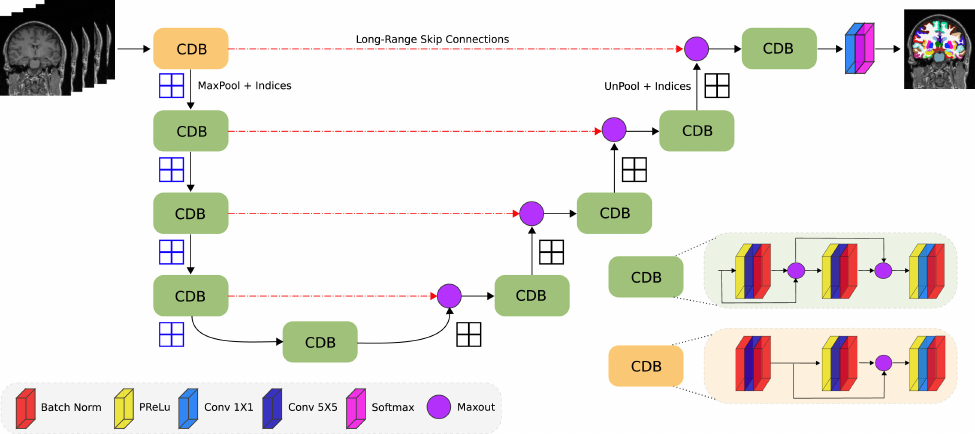}
  \caption{Illustration of FastSurfer's architecture. The CNN consists of four
    competitive dense blocks (CDB) in the encoder and decoder part, separated by a
    bottleneck layer. Figure reproduced from~\citep{henschel2020fastsurfer}.}
  \label{fig:fastsurfer}
\end{figure}

FastSurfer~\citep{henschel2020fastsurfer} is a CNN model that performs
whole-brain segmentation, cortical surface reconstruction, fast spherical
mapping, and cortical thickness analysis from anatomical MRI. The FastSurfer CNN is inspired by the QuickNAT
model~\citep{roy2019quicknat}, which is composed of three 2D fully
convolutional neural networks---each associated with a different 2D slice
orientation---that each have the same encoder/decoder U-net architecture with
skip connections, unpooling layers and dense connections as QuickNAT. A diagram
of the model's architecture is available in the Appendix
Figure~\ref{fig:fastsurfer}. We focus exclusively on the task of whole-brain
segmentation, defined as voxel-wise anatomical labeling of brain regions. This
segmentation step is entirely performed by the CNN, without surface
reconstruction, and uses the pre-trained FastSurfer model (v2.1.0) available on
GitHub~\citep{fastsurfer-github}. FastSurfer has demonstrated high accuracy,
strong generalization to unseen datasets, and high test-retest reliability.
This model serves as an ideal benchmark for studying numerical variability in
high-dimensional medical imaging tasks due to its scientific relevance and
architectural complexity. In our experiments, we applied FastSurfer to segment
five subjects from the CoRR dataset~\citep{zuo2014open}.
By comparing Fuzzy
PyTorch and Verrou instrumentation on FastSurfer inference, we aim to partially
replicate previous findings on numerical uncertainty in
FastSurfer~\citep{pepe2023numerical}, thereby validating the accuracy,
reliability, and applicability of Fuzzy PyTorch to realistic large-scale segmentation tasks.
For FastSurfer, we cannot compute significant bits as its segmentations are composed of categorical integer labels. Therefore, we compute the minimum Sørensen-Dice score across pairs of stochastic arithmetic runs. The Sørensen-Dice score measures the overlap between two segmentation results and is commonly used to quantify similarity between labeled regions in medical imaging. Further details on the Sørensen-Dice score are provided in Appendix~\ref{subsec:dice_scores}.

\section{Sørensen-Dice Scores}
\label{subsec:dice_scores}
The neuroimaging use case, the FastSurfer convolutional neural network (CNN), produces brain segmentations from anatomical Magnetic Resonance
Images (MRI), which label different brain regions. In order to evaluate brain segmentations, we cannot use the significant bits
metric. Segmentation tools like FastSurfer generate categorical
variables encoded as integers to represent segmentation labels, even though
they rely on floating-point operations. Therefore, the significant bits metric
cannot be applied as it is only useful for programs that produce floating-point
outputs.

To assess the impact of numerical perturbations on segmentation stability, we
compute the minimum Sørensen-Dice score across pairs of stochastic arithmetic runs. The
Sørensen-Dice score measures the overlap between two segmentation results and
is commonly used to quantify similarity between labeled regions in medical
imaging. For each subject, we run FastSurfer multiple times with SA-enabled
perturbations, producing $N$ different segmentations. Each segmentation output
assigns a class label to every voxel in the brain MRI. Let $S_i$ and $S_j$
represent the set of voxels assigned to a specific brain region in the $i$-th
and $j$-th stochastic arithmetic iterations, respectively. The Sørensen-Dice score between these
two segmentations is given by:


\[
  \text{min Sørensen-Dice Score} = \min_{\substack{i, j \in \{1, \dots, N\} \\ i \neq j}} \left( \frac{2 \cdot |S_i \cap S_j|}{|S_i| + |S_j|} \right)
\]

where $|S_i \cap S_j|$ is the number of overlapping voxels classified as part
of the same region in both segmentations, and $|S_i|$ and $|S_j|$ are the total
number of voxels assigned to that region in each iteration.

\section{Statistical Analysis of Harmonic Series Variance Homogeneity}

\begin{table}[H]
  \begin{center}
    \begin{tabular}{|l|c|c|c|c|}
      \hline
      \multicolumn{5}{|c|}{\textbf{Descriptive Statistics by Method}}                                                                  \\
      \hline
      \textbf{Method} & \textbf{Mean Std. Dev.} & \textbf{95\% CI}              & \textbf{F-statistic vs. PRISM SR} & \textbf{p-value} \\
      \hline
      PRISM SR        & $1.65 \times 10^{-4}$   & $[1.08, 2.23] \times 10^{-4}$ & ---                               & ---              \\
      MCA RR          & $2.41 \times 10^{-4}$   & $[1.37, 3.45] \times 10^{-4}$ & $F = 1.58$                        & $p = 0.210$      \\
      Verrou SR       & $1.48 \times 10^{-4}$   & $[1.00, 1.96] \times 10^{-4}$ & $F = 0.21$                        & $p = 0.651$      \\
      FM SR           & $2.10 \times 10^{-4}$   & $[1.27, 2.94] \times 10^{-4}$ & $F = 0.78$                        & $p = 0.380$      \\
      \hline
    \end{tabular}
  \end{center}
  \caption{Statistical analysis of variance homogeneity across stochastic rounding methods in harmonic series computation.
    Levene's test confirms homogeneity of variances across stochastic rounding implementations ($F=1.26$, $p=0.29$),
    supporting the validity of comparative analyses. All pairwise F-tests compare against PRISM SR as the reference method.
    Tests performed at $\alpha = 0.05$ significance level.}
  \label{tab:harmonic_stats}
\end{table}

\end{document}